\documentclass[runningheads]{llncs}
\usepackage[T1]{fontenc}
\usepackage{graphicx}
\usepackage{bm}
\usepackage{amssymb}
\usepackage{amsmath}
\usepackage{enumitem}
\usepackage{booktabs}
\usepackage{graphicx}
\usepackage{subcaption}
\usepackage{xcolor}
\begin{document}
\title{Dual-Agent Multiple-Model Reinforcement Learning for Event-Triggered Human-Robot Co-Adaptation in Decoupled Task Spaces}
\titlerunning{Dual-Agent MRL for HR Co-Adaptation}
%
\author{Yaqi Li\inst{2}\textsuperscript{\dag} \and Zhengqi Han\inst{2}\textsuperscript{\dag} \and Huifang Liu\inst{2}\textsuperscript{\dag} \and Steven W. Su\inst{1,2}\thanks{Corresponding author: steven.su@uts.edu.au}
}
\authorrunning{Y. Li et al.}
%
\institute{ College of Medical information and Artificial intelligence, Shandong First Medical University (Shandong Academy of Medical Science) 6699 Qingdao Road, Jinan, Shandong, China \and Faculty of Engineering and IT, University of Technology Sydney, Australia }
\maketitle 
\footnotetext{\textsuperscript{\dag} These authors contributed equally to this work.}
\begin{abstract}
	This paper presents a shared-control rehabilitation policy for a custom 6-degree-of-freedom (6-DoF) upper-limb robot that decomposes complex reaching tasks into decoupled spatial axes. The patient governs the primary reaching direction using binary commands, while the robot autonomously manages orthogonal corrective motions. Because traditional fixed-frequency control often induces trajectory oscillations due to variable inverse-kinematics execution times, an event-driven progression strategy is proposed. This architecture triggers subsequent control actions only when the end-effector enters an admission sphere centred on the immediate target waypoint, and was validated in a semi-virtual setup linking a physical pressure sensor to a MuJoCo simulation. To optimise human--robot co-adaptation safely and efficiently, this study introduces Dual Agent Multiple Model Reinforcement Learning (DAMMRL). This framework discretises decision characteristics: the human agent selects the admission sphere radius to reflect their inherent speed--accuracy trade-off, while the robot agent dynamically adjusts its 3D Cartesian step magnitudes to complement the user's cognitive state. Trained in simulation and deployed across mixed environments, this event-triggered DAMMRL approach effectively suppresses waypoint chatter, balances spatial precision with temporal efficiency, and significantly improves success rates in object acquisition tasks.
	
\end{abstract}

\section{Introduction}

Robot-assisted upper-limb rehabilitation aims to deliver high-dosage, task-oriented practice while reducing clinician burden and improving repeatability. Decades of device development---from end-effector to exoskeleton systems---have shown that robotic training can standardise repetition and personalise assistance, yet real-world performance still hinges on how effectively human intent is captured and fused with robot autonomy \cite{Maciejasz2014survey}. Two recurring bottlenecks are: (i) intent-decoding pipelines that are accurate yet light enough for real-time control, and (ii) control policies that avoid oscillations and indecision around task waypoints, especially near targets where fixed-frequency updates can create chatter due to variable execution times in inverse kinematics.

On the sensing side, wearable interfaces have matured from low-channel IMU/EMG classifiers to rich, high-density arrays, enabling low-latency decoding of discrete motor intent in impaired users \cite{Meyers2024stroke,Tacca2024hdEMG,Anselmino2024gait}. For robot policy design, shared/autonomous strategies that keep the human ``in the loop'' while the robot resolves kinematic and grasp constraints have improved task success and perceived controllability in assistive manipulation \cite{Bowman2024shared,Styler2025shared}. In rehabilitation-specific controllers, assist-as-needed (AAN) schemes adapt the assistance level to the user's performance to promote engagement and motor relearning \cite{Lu2024aan,Leerskov2024hybrid}. A parallel and complementary thread in control theory argues for \emph{event-triggered} updates---progressing only when measurable conditions are met---thereby reducing needless actions and mitigating oscillations that often arise under fixed-rate sampling \cite{Heemels2012event}.

Building on these trends and our initial MuJoCo simulations, we propose an \emph{axial decomposition} policy for a custom 6-DoF upper-limb robot. This approach assigns the user a binary decision on the primary reaching axis (e.g., up/down inferred via IMU, EMG, or EEG), while the robot autonomously selects corrective spatial motions on the orthogonal axes. Each Cartesian step is mapped to joint space using damped least-squares inverse kinematics (IK) with dynamics-consistent tracking and impedance shaping for comfort and safety \cite{Chiaverini1994DLS,Deo1995DLS,Hogan1985imp}. To address the waypoint-chatter problem, step progression is governed by an \emph{event-driven strategy} rather than a fixed timer. Actions are updated only when the robot's end-effector reaches an admission sphere centred on the immediate target waypoint, which empirically suppresses back-and-forth oscillations.

To address inter-individual variability without incurring heavy, continuous online adaptation, we introduce Dual Agent Multiple Model Reinforcement Learning (DAMMRL). This DQN-based discrete co-adaptation scheme explicitly models the human as $\mathrm{Agent}_0$ and the robot as $\mathrm{Agent}_1$. The system quantises decision characteristics into a finite multi-model set $\mathcal{M}=\{\mathcal{M}_{i,j}\}$. Here, the index $i \in \{1, 2\}$ represents the human agent's selected admission sphere radius $\varepsilon \in \{E_{\text{big}}, E_{\text{small}}\}$, which corresponds to their inherent speed--accuracy trade-off. The index $j \in \{1, \dots, 8\}$ enumerates the $2^3 = 8$ possible combinations of the robot agent's 3D step magnitude vector $\bm{\delta} = [\delta_x, \delta_y, \delta_z]^\top$, where each axial component independently takes a small or large step value (e.g., $\delta_x \in \{x_s, x_b\}$). The DAMMRL curriculum selects the optimal multi-model matching $\mathcal{M}_{i,j}$ to balance spatial accuracy and time efficiency. The curriculum proceeds sequentially across three environments: 
(1) Virtual (both agents simulated in MuJoCo; RL training executed), $\rightarrow$ 
(2) Semi-virtual (real human governing input via a pressure sensor with the virtual MuJoCo robot; model refinement), $\rightarrow$ 
(3) Real (both agents physical; fully real-world deployment).

This finite-model co-adaptation treats the human as a learning agent and optimises collaboration rather than unilateral assistance \cite{Guo2024coopMDP,An2024DAMMRL}.

Building on the design outlined above, the major contributions of this study can be summarised as follows:
\begin{itemize}
	\item[(i)] an axial human--robot role allocation that reduces intent decoding to robust binary decisions while preserving user agency over task progress;
	\item[(ii)] an event-driven progression criterion, utilising an admission sphere, that suppresses waypoint oscillations commonly observed under fixed-rate updates;
	\item[(iii)] a DQN-based DAMMRL framework that maps fixed-length Cartesian micro-steps to six-joint trajectories via inverse dynamics, discretely matching error sphere radii with accuracy demands; and
	\item[(iv)] a staged, finite-model co-adaptation pipeline progressing seamlessly from MuJoCo simulations to semi-virtual and ultimately physical environments, thereby simplifying on-hardware tuning and deployment.
\end{itemize}

\section{Related Work}
Related areas include wearable-sensor intent decoding for discrete commands, shared autonomy in Cartesian spaces, co-adaptive control for rehabilitation, and event-triggered control to reduce chatter and resource use. Our approach integrates these strands into a practical policy combining axial binary intent, autonomous orthogonal adjustments, inverse-dynamics mapping, and a staged Dual Agent Multiple Model Reinforcement Learning (DAMMRL) pipeline.

\section{Proposed System and Methodology}
\subsection{System Overview: Decoupled Task Spaces}

\begin{figure}[htbp]
	\centering
	\begin{subfigure}[b]{0.48\textwidth}
		\centering
		\includegraphics[width=\textwidth]{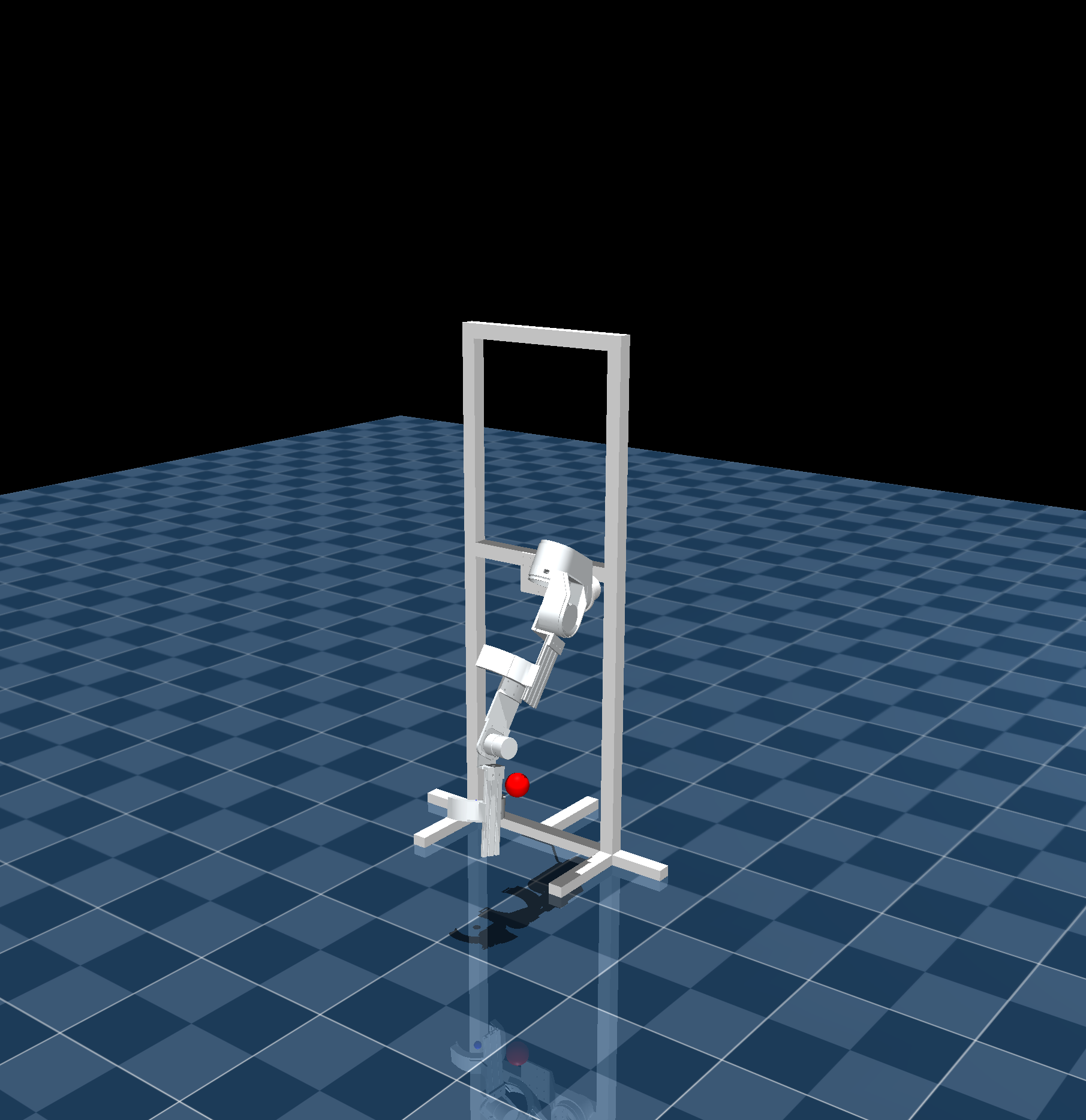}
		\caption{MuJoCo simulation environment}
		\label{fig:virtual-env}
	\end{subfigure}
	\hfill
	\begin{subfigure}[b]{0.32\textwidth}
		\centering
		\includegraphics[width=\textwidth]{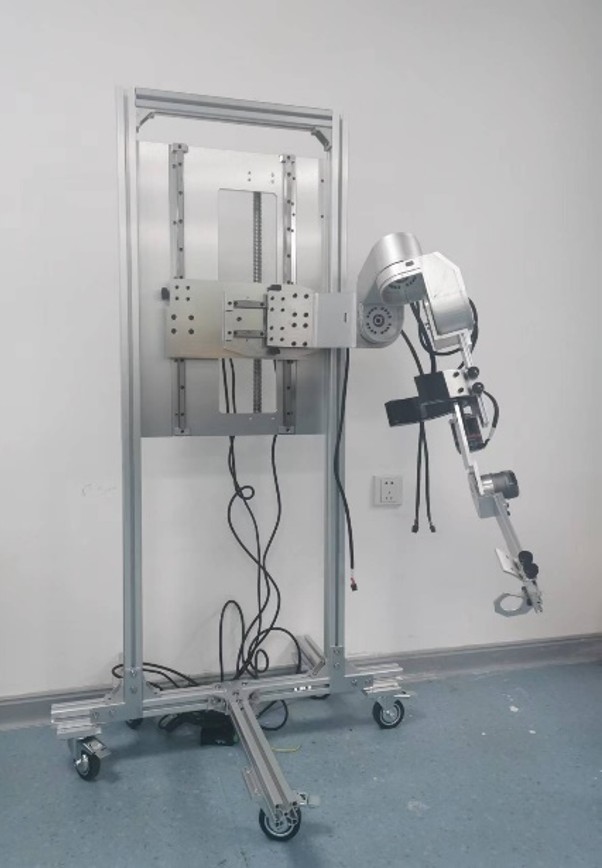}
		\caption{Physical 6-DoF manipulator setup}
		\label{fig:real-env}
	\end{subfigure}
	
	\caption{Experimental setups for the 6-DoF robotic manipulator, illustrating (a) the virtual training environment and (b) the real-world physical deployment.}
	\label{fig:experimental-setups}
\end{figure}

\paragraph{\textbf{Robot and Frames}}
We consider a custom 6-DoF serial manipulator with joint coordinates $\bm q \in \mathbb{R}^6$ and an end-effector position $\bm x \in \mathbb{R}^3$ defined within a task frame. The primary objective is to move the end-effector from a starting position $\bm x_s$ to a goal position $\bm x_g$ (the application scenario involves a patient pushing a button with the assistance of the robot). The experimental validation progresses through staged environments: the virtual environment, simulated in MuJoCo, is illustrated in Fig. \ref{fig:virtual-env}, while the fully physical robot setup is shown in Fig. \ref{fig:real-env}.

\paragraph{\textbf{Human Sensing and Binary Intent}}
Wearable physiological sensors (e.g., IMU, EMG, EEG) and direct-control interfaces (e.g., mechanical switches or pressure sensors) provide a discrete binary decision $u_h\in\{-1,+1\}$ governing movement along the primary reaching axis (e.g., the $z$-axis, defined as up/down). A two-class classifier $\pi_h(\cdot)$ processes short temporal windows to output a directional command and its associated confidence. This forms the basis of the human agent ($\mathrm{Agent}_0$) input.

\paragraph{\textbf{Robot Autonomy on Orthogonal Axes and Step Sizing}}
Given the spatial error $\bm e=\bm x_g-\bm x$, the robot autonomous agent ($\mathrm{Agent}_1$) selects the directional signs of the Cartesian steps along the orthogonal axes ($y$ and $z$) to minimise lateral and vertical deviations, whilst the human agent advances the system along the primary $x$-axis. Additionally, the robot agent determines the precise moving distances (step lengths) for all three spatial directions ($x$, $y$, and $z$). This dynamic sizing corresponds directly to the Cartesian step parameters ($[x_s, x_b, y_s, y_b, z_s, z_b]$) defined within the robot's submodel, allowing for adaptive spatial scaling during the reaching task.

\subsection{Event-Driven Progression and Admission Spheres}
\label{subsec:trigger}
Traditional discrete-time control strategies with fixed-time sampling led to back-and-forth oscillations (chatter) near subgoals, primarily because mapping Cartesian movements to six motor rotations via inverse kinematics does not consume a uniform time budget. To mitigate this, we propose an event-driven progression strategy. We trigger the subsequent control action only when the end-effector enters an admission sphere centred on the immediate target waypoint, and the system exhibits energetic convergence:
\begin{align}
	\text{Trigger:}\quad
	\left \|\bm x - \bm x^{(m)} \right \| \le \varepsilon
	\ \wedge\
	\dot{V} \le 0,
\end{align}
where $\bm x^{(m)}$ is the current subgoal and $\varepsilon$ represents the radius of the admission sphere (which directly corresponds to the error sphere radii $[E_{\text{big}}, E_{\text{small}}]$ defined in the human submodel). The term $V=\tfrac12 \bm e^\top \mathbf W \bm e$ acts as a Lyapunov surrogate with a positive definite weight matrix $\mathbf W\succ0$. This event-triggered architecture ensures sufficient spatial settling before advancing, suppressing oscillations and improving success rates in object acquisition tasks.

\subsection{Axial Decomposition and Dynamics-Consistent Control}
\paragraph{\textbf{Problem Motivation}}
Relying purely on position control (e.g., joint-space PD control or stepping directly to target joint angles) for waypoint navigation often leads to overshoot, oscillation, and rebound when approaching sub-goals. This instability primarily arises from unmodelled dynamic terms---such as inertia, Coriolis and centrifugal forces, and gravity---which amplify momentum during rapid motions or posture coupling. Furthermore, the lack of adjustable compliance between the end-effector and the user (or environment) makes the force--displacement relationship uncontrollable, negatively affecting comfort and stability.

To address these challenges and ensure smooth energy exchange, we propose a ``kinematics--dynamics consistent'' control strategy. This framework first decomposes the reaching motion into discrete Cartesian micro-steps within the decoupled task space. These dynamically sized steps are then mapped to the joint space via inverse kinematics. Finally, inverse dynamics and impedance shaping are employed to generate the appropriate joint torques, enabling smooth and predictable physical interaction.

\paragraph{\textbf{Dynamic Axial Step Sizing}}
The foundation of our approach relies on decomposing the motion into discrete Cartesian steps where the step lengths are actively assigned by the DAMMRL policy. Rather than relying on a static, fixed-length step, the Cartesian displacement vector $\Delta \bm x$ is formulated as:
\begin{align}
	\Delta \bm x & = 
	\begin{bmatrix}
		\Delta_x \\ \Delta_y \\ \Delta_z
	\end{bmatrix} = 
	\begin{bmatrix}
		u_h \delta_x \\ \operatorname{sgn}(e_y) \delta_y \\ \operatorname{sgn}(e_z) \delta_z
	\end{bmatrix},
	\label{eq:cart_step}
\end{align}
where $u_h \in \{-1, +1\}$ is the binary intent on the primary $x$-axis, and $\operatorname{sgn}(\cdot)$ determines the direction along the orthogonal $y$ and $z$ axes based on the spatial error $\bm e$. Crucially, the magnitude parameters $\delta_x \in \{x_s, x_b\}$, $\delta_y \in \{y_s, y_b\}$, and $\delta_z \in \{z_s, z_b\}$ are the specific moving distances determined autonomously by the robot agent ($\mathrm{Agent}_1$) at each step, ensuring adaptive spatial scaling as the end-effector approaches the target.

\paragraph{\textbf{Kinematic Layer: Numerical Inverse Kinematics}}
Given the current end-effector position $\mathbf{x}_k$ and the generated Cartesian micro-step $\Delta \mathbf{x} \in \mathbb{R}^3$, the immediate spatial target is defined as $\mathbf{x}_{k+1} = \mathbf{x}_k + \Delta \mathbf{x}$. To map this task-space target into the joint space, we utilise an optimisation-based numerical inverse kinematics approach rather than a Jacobian-based pseudo-inverse method. Specifically, the inverse kinematics are resolved using the \texttt{ikpy} library, which formulates the IK resolution as a bounded optimisation problem to minimise the Cartesian error between the manipulator's forward kinematics and the target $\mathbf{x}_{k+1}$. Under the event-triggered micro-step strategy, this solver is queried once after each event trigger, yielding the reference joint trajectory seed $\mathbf{q}_{k+1}$ for the subsequent dynamic control layer.

\paragraph{\textbf{Dynamic Layer: Analytical Acceleration and Inverse Dynamics}}
At the dynamic layer, the analytical acceleration method is employed to map the desired end-effector acceleration into the joint space, ensuring explicit compensation for the manipulator's nonlinear dynamics. The rigid-body dynamics in the joint space are expressed as:
\begin{equation}
	\mathbf{M}(\mathbf{q})\ddot{\mathbf{q}} + \mathbf{C}(\mathbf{q},\dot{\mathbf{q}})\dot{\mathbf{q}} + \mathbf{g}(\mathbf{q}) \;=\; \boldsymbol{\tau},
	\label{eq:manipulator}
\end{equation}
where $\mathbf{q} \in \mathbb{R}^6$ represents the joint angles, $\mathbf{M}(\mathbf{q})$ is the symmetric positive-definite inertia matrix, $\mathbf{C}(\mathbf{q},\dot{\mathbf{q}})$ captures the Coriolis and centrifugal forces, and $\mathbf{g}(\mathbf{q})$ denotes the gravity vector. The kinematic relationship at the acceleration level satisfies:
\begin{equation}
	\ddot{\mathbf{x}} \;=\; \mathbf{J}(\mathbf{q})\ddot{\mathbf{q}} + \dot{\mathbf{J}}(\mathbf{q},\dot{\mathbf{q}})\dot{\mathbf{q}},
\end{equation}
where $\mathbf{J}(\mathbf{q})$ is the analytical Jacobian matrix.

Given a desired task-space acceleration $\ddot{\mathbf{x}}_d$ (which embeds the event-triggered spatial step commands), the required joint acceleration $\ddot{\mathbf{q}}_d$ is resolved via:
\begin{equation}
	\ddot{\mathbf{q}}_d \;=\; \mathbf{J}^{\#}\big(\ddot{\mathbf{x}}_d - \dot{\mathbf{J}}\dot{\mathbf{q}}\big) 
	\;+\; \big(\mathbf{I}-\mathbf{J}^{\#}\mathbf{J}\big)\mathbf{z},
\end{equation}
where $\mathbf{J}^{\#}$ is the Moore-Penrose pseudo-inverse of the Jacobian matrix, and $\mathbf{z}$ is an arbitrary joint-acceleration vector projected into the null space. This null-space term can be utilised for secondary objectives without affecting the primary task, such as joint limit avoidance or energy minimisation.

After resolving the desired joint trajectories via inverse kinematics, a standard Computed Torque Control (CTC) law is employed at the dynamic layer. This leverages an inverse-dynamics feedforward model alongside a joint-level PD feedback loop to compensate for the manipulator's inertia, Coriolis forces, and gravity, ensuring accurate and stable torque execution.

\subsection{Dual Agent Multiple Model Reinforcement Learning (DAMMRL) Co-Adaptation}
\paragraph{\textbf{Discrete Human and Robot Models}}
The DAMMRL framework consists of two parallel, co-adaptive decision-making agents. Both agents share a common continuous state space, where the fundamental observation 
$\bm{s} = (\bm{x}, \bm{x}_g)$ 
comprises the current end-effector position and the final target position. To manage individual variability without continuous online adaptation, the system quantises human and robot capabilities into a finite set of multi-model combinations $\mathcal{M}=\{\mathcal{M}_{i,j}\}$. Here, $i \in \{1, 2\}$ denotes the human's discrete cognitive state dictated by their selected admission sphere, and $j \in \{1, \dots, 8\}$ enumerates the robot's active 3D step magnitude vector.

\begin{itemize}
\item \textbf{Human Agent ($\mathrm{Agent}_0$):} The human governs the primary reaching axis through a dual decision-making process. First, they provide a discrete directional command $u_h \in \{-1, +1\}$ (e.g., up/down along the $z$-axis). Second, they select the required proximity to the subsequent step destination by determining the admission sphere radius $\varepsilon \in \{E_{\text{big}}, E_{\text{small}}\}$. To account for the inherent speed--accuracy trade-off and cognitive errors present in real-world human--robot interaction, the simulated human is modelled with a stochastic policy based on empirical observations. Specifically, when the human selects $E_{\text{big}}$, they must make decisions more rapidly, which induces a higher error rate of approximately 20\% (i.e., an accuracy of 80\%, corresponding to the $A_{\text{low}}$ demand). Conversely, selecting $E_{\text{small}}$ allows for more careful, deliberate decision-making, yielding a lower error rate of 10\% (i.e., an accuracy of 90\%, corresponding to the $A_{\text{high}}$ demand). 

This selection dynamically adjusts the event-trigger threshold, explicitly linking the user's cognitive load and decision frequency to the system's spatial accuracy requirements.

	\item \textbf{Robot Agent ($\mathrm{Agent}_1$):} Operating in tandem with the human's temporal and spatial selections, the robot governs the orthogonal corrective motions and the overall step magnitudes. Its discrete action space is formulated as $\mathcal{A}_m \in \{\delta_{\text{small}}, \delta_{\text{large}}\}^3$. Rather than relying on a static, global step size, the model outputs independent magnitude actions for the $x$, $y$, and $z$ axes at each event trigger, dynamically selecting from the parameter set $[x_s, x_b, y_s, y_b, z_s, z_b]$. By autonomously adjusting these axial step lengths, the robot directly compensates for or complements the user's chosen admission sphere ($\varepsilon$) and its corresponding error rate. For instance, the robot policy learns to pair specific step sizes with the human's $E_{\text{big}}$ (higher speed, 20\% error) versus $E_{\text{small}}$ (lower speed, 10\% error) states.

	This multi-model matching ensures that the robot can adaptively optimise both spatial reaching efficiency and final waypoint precision, perfectly balancing the human's speed--accuracy trade-off during the co-adaptive task.
\end{itemize}

\paragraph{\textbf{Training Curriculum: Virtual to Semi-Virtual to Real}}
To ensure safe and efficient deployment, the finite-model co-adaptation pipeline proceeds sequentially across three staged environments:
\begin{enumerate}[leftmargin=1.15em]
	\item \textbf{Virtual (Sim--Sim):} Both the human and robot agents are fully simulated within the MuJoCo environment. The DAMMRL algorithm executes extensive RL training to explore the finite model set $\{\mathcal{M}_{i,j}\}$ and learn optimal matches that balance speed, accuracy, and comfort.
	\item \textbf{Semi-Virtual (Human--Sim):} A real human participant interacts with the virtual MuJoCo robot via a physical interface. In our specific implementation, the human governs the $x$-axis via a pressure sensor, and the directional commands are executed in the virtual environment via wireless serial port communication. This stage refines the selected models by estimating human decision frequency and accuracy online.
	\item \textbf{Real (Human--Real):} The optimally matched multi-model combinations are deployed onto the physical 6-DoF hardware. The posterior distribution over $\{\mathcal{M}_{i,j}\}$ is updated based on real-world task performance and user comfort. If task targets are unmet, the system iterates and selects alternative matched models. \bf{Due to time constraints, this test has not yet been implemented but is planned for a future study.}
\end{enumerate}


%


\paragraph{\textbf{State, Actions, and Reward Formulation}}
The augmented RL state vector is $\bm{s} = [\bm{x};\, \bm{x_g};\, \mathcal{M}_{i,j}]$. The composite actions include the dynamic axial step selections by the robot and the directional commands by the human. The system objective is captured by a composite reward function designed to penalise tracking error, excessive execution time, mechanical effort, and waypoint chatter, whilst rewarding successful target acquisition:
\begin{align}
	r &= \alpha \frac{1}{\left \| \bm{e} \right \|^2+0.5}  - \beta \, t_{\text{step}} - \gamma \, E_{\text{effort}} 
	- \eta \, N_{\text{osc}} + \rho \, R_{\text{success}},
\end{align}
where $\alpha, \beta, \gamma, \eta$, and $\rho$ are positive scalar weights.

\subsection{Overall Algorithm}
\textbf{Algorithm 1: Event-Driven Axial Co-Adaptation via DAMMRL}
\begin{enumerate}[leftmargin=1.15em]
	\item \textbf{Initialise:} Manipulator joint configuration $\bm{q}_0$, measure initial position $\bm{x}_0$, set subgoal index $m=1$, and select the initial multi-model match $(i,j)$ from $\mathcal{M}$.
	\item \textbf{Loop} until final target convergence $\left \| \bm{x} - \bm{x}_g \right \| \le \varepsilon_g$:
	\begin{enumerate}[leftmargin=1em]
		\item Human agent ($\mathrm{Agent}_0$) executes a dual decision: outputting the primary axis directional command $u_h\in\{-1,+1\}$ and selecting the admission sphere radius $\varepsilon \in \{E_{\text{big}}, E_{\text{small}}\}$ governed by their speed--accuracy trade-off.
		\item Robot agent ($\mathrm{Agent}_1$) evaluates the current spatial error $\bm{e}$ in tandem with the human's selected $\varepsilon$. It autonomously selects orthogonal directional signs ($\operatorname{sgn}(e_y), \operatorname{sgn}(e_z)$) and dynamic step magnitudes ($\delta_x, \delta_y, \delta_z$) from the set $[x_s, x_b, y_s, y_b, z_s, z_b]$ to optimise reaching efficiency and compensate for the human's expected error rate.
		\item Formulate the Cartesian micro-step $\Delta\bm{x}$ via \eqref{eq:cart_step}.
		\item Resolve the new target position ($\bm{x} + \Delta\bm{x}$) into the joint-space configuration $\mathbf{q}_{k+1}$ using an optimisation-based numerical inverse kinematics solver (\texttt{ikpy}).
		\item \textbf{Event trigger progression:} Advance the subgoal index $m\leftarrow m+1$ \emph{only if} the end-effector enters the dynamically sized admission sphere ($\| \bm{x} - \bm{x}^{(m)} \| \le \varepsilon$) defined by the human in step (a), and the Lyapunov energy surrogate satisfies $\dot{V} \le 0$. Otherwise, hold the current target position.
		\item Update the model posterior probability using online estimates of the human's effective error rate and decision frequency, alongside the observed execution time of the robot.
	\end{enumerate}
\end{enumerate}

\section{Experiments and Results}
\subsection{Experimental Setups}

To validate the proposed Dual Agent Multiple Model Reinforcement Learning (DAMMRL) framework and the event-driven control architecture, the experimental validation was conducted across three progressively complex stages:

\begin{itemize}
	\item \textbf{S1 Virtual (Sim--Sim):} A fully virtual training environment simulated in MuJoCo, as illustrated in Fig. \ref{fig:virtual-env}. In this stage, both the human and robot agents are simulated. The environment incorporates randomised manipulator kinematics and dynamics, as well as simulated sensor noise, to ensure robust policy learning. The simulated human agent executes decisions with stochastic error rates corresponding to their selected speed--accuracy trade-off ($E_{\text{big}}$ versus $E_{\text{small}}$).
	
	\item \textbf{S2 Semi-Virtual (Human--Sim):} A hybrid experimental setup involving a real human participant interacting with the virtual MuJoCo robot. In this stage, the human participant governs the reaching axis using a physical pressure sensor, with directional commands transmitted to the simulation via wireless serial port communication. This stage refines the policies by estimating the user's actual decision frequency and accuracy online, narrowing down the optimal candidate models. Healthy participants were recruited for this phase under Shandong First Medical University (SDFMU) Ethics Approval ID: R202503040145.
	
	\item \textbf{S3 Real (Human--Real):} The fully physical environment utilising our custom 6-DoF upper-limb robotic manipulator, as shown in Fig. \ref{fig:real-env}. The optimally matched multi-model combinations derived from S1 and refined in S2 are deployed directly onto the hardware. This stage physically validates the event-driven progression, the dynamically sized Cartesian micro-steps, and the pure inverse-dynamics feedforward control during real-world collaborative task execution. \footnote{Due to time constraints, this test has not yet been implemented but is planned for a future study.}
\end{itemize}

\subsection{Evaluation Metrics}

To comprehensively assess the performance of the proposed DAMMRL framework and the event-driven progression strategy, the system's efficacy was evaluated across several quantitative dimensions. These metrics are specifically aligned with the composite reward function and the core objectives of rehabilitation robotics:

\begin{itemize}
	\item \textbf{Task Efficacy and Temporal Efficiency:}
	\begin{itemize}
		\item \textit{Success Rate ($R_{\text{success}}$):} The percentage of trials where the end-effector successfully reached the final target position within the acceptable global tolerance ($\| \bm{x} - \bm{x}_g \| \le \varepsilon_g$) without violating joint limits or safety constraints.
		\item \textit{Execution Time ($t_{\text{total}}$):} The total time elapsed from the initiation of the task ($\bm{x}_s$) to final target convergence. We also evaluated the average micro-step duration ($t_{\text{step}}$) to assess the temporal impact of the dynamic axial step sizing.
	\end{itemize}
	
	\item \textbf{Spatial Accuracy and Kinematic Stability:}
	\begin{itemize}
		\item \textit{Final Positioning Error ($\| \bm{e} \|$):} The absolute Euclidean distance between the end-effector's final resting position and the desired goal $\bm{x}_g$.
		\item \textit{Waypoint Oscillation Count ($N_{\text{osc}}$):} A critical metric for validating the event-driven strategy. It quantifies the number of direction reversals or "chatter" occurrences near the admission spheres ($\varepsilon$), comparing the proposed event-triggered approach against traditional fixed-time sampling.
		\item \textit{Trajectory Smoothness:} Evaluated via the integrated squared jerk of the end-effector trajectory, ensuring that the pure inverse-dynamics feedforward control provides a comfortable, dynamically consistent motion without abrupt accelerations.
	\end{itemize}
	
	\item \textbf{Human--Robot Co-Adaptation and Cognitive Load:}
	\begin{itemize}
		\item \textit{Empirical Error Rate:} Monitored specifically during the S2 (Semi-Virtual) and S3 (Real) stages to validate the assumed stochastic policy. This measures the frequency of incorrect directional commands $u_h$ relative to the human's selected admission sphere radius ($E_{\text{big}}$ vs. $E_{\text{small}}$), confirming the speed--accuracy trade-off.
		\item \textit{Multi-Model Convergence:} The number of iterations required for the DAMMRL algorithm to converge on the optimal discrete parameter set $(i,j) \in \mathcal{M}$ that best balances the human's specific decision frequency with the robot's step magnitudes ($\delta_x, \delta_y, \delta_z$).
	\end{itemize}
\end{itemize}

\subsection{Comparison Study}

Due to development time constraints, the comprehensive comparative evaluation of the control policies was executed exclusively within the fully virtual MuJoCo environment (S1). The semi-virtual environment (S2) was successfully implemented to validate the human--robot interface, specifically utilising a physical pressure-pushing actuator to govern the user's primary reaching axis and admission sphere selection. However, the rigorous quantitative benchmarking presented below relies on the S1 simulations to isolate the effects of the control algorithms. 

To evaluate the efficacy of the proposed architecture, we compared four distinct experimental configurations:

\begin{enumerate}[label=\textbf{\arabic*.}, leftmargin=1.5em]
	\item \textbf{Fixed-Frequency Control (Baseline):} A traditional discrete-time strategy where Cartesian micro-steps are updated at a constant, predefined sampling rate. This baseline ignores the end-effector's spatial proximity to the subgoal before initiating the next step, serving to illustrate the impact of inverse kinematics time-budget variations.
	
	\item \textbf{Fixed-Model Event-Driven Control:} This configuration employs the proposed admission sphere ($\varepsilon$) progression strategy, triggering steps only when spatial and energetic criteria are met ($\| \bm{x} - \bm{x}^{(m)} \| \le \varepsilon \wedge \dot{V} \le 0$). However, it uses static, predefined step lengths ($\delta_x, \delta_y, \delta_z$) without the co-adaptive DAMMRL framework, demonstrating the isolated benefit of the event trigger.
	
	\item \textbf{Proposed Event-Driven DAMMRL (Reward 1 - Accuracy Focus):} Our proposed dual-agent RL framework trained with a restricted reward function. This formulation heavily incentivises spatial accuracy by mapping the spatial error ($\| \bm{e} \|$) to a bounded positive reward signal via the term $\alpha \frac{1}{\| \bm{e} \|^2+0.5}$. It simultaneously penalises waypoint oscillations ($N_{\text{osc}}$) while explicitly omitting the temporal penalty ($\beta = 0$). This configuration isolates the algorithm's capacity to maximise spatial precision and trajectory stability, completely irrespective of the overall execution time.
	
	\item \textbf{Proposed Event-Driven DAMMRL (Reward 2 - Speed and Accuracy Balance):} The complete, fully formulated architecture utilising the comprehensive composite reward function. This model trains the robot agent to dynamically adjust axial step magnitudes to optimally balance the human's selected speed--accuracy trade-off ($E_{\text{big}}$ vs. $E_{\text{small}}$), ensuring both spatial precision and temporal efficiency.
\end{enumerate}

\paragraph{\textbf{Expected Comparative Outcomes}}
While the Fixed-Frequency baseline (Experiment 1) typically suffers from severe waypoint chatter and oscillations due to mismatched dynamic execution times, introducing the event trigger in Experiment 2 effectively suppresses this instability. However, without adaptive step sizing, the Fixed-Model approach struggles to optimise reaching speed. 

Under the DAMMRL framework, Experiment 3 (Reward 1) is expected to demonstrate exceptional spatial accuracy and zero chatter, but at the cost of prolonged execution times, as the robot agent defaults to overly cautious, small step sizes ($\delta_{\text{small}}$). Finally, Experiment 4 (Reward 2) represents the optimal co-adaptive solution. By explicitly penalising both time and error, the robot agent dynamically shifts between $\delta_{\text{large}}$ and $\delta_{\text{small}}$ to match the human's cognitive state, achieving a smooth, oscillation-free trajectory that minimises both effort and task duration.

\subsection{Results}

Overall, the experimental results demonstrate a progressive performance improvement from the fixed-frequency baseline to fully proposed DAMMRL architecture. The incorporation of event-triggered mechanisms significantly enhances spatial stability, while the introduction of dual-agent reinforcement learning further improves adaptability and efficiency.

\begin{figure}[htbp]
	\centering
	\includegraphics[width = 0.8\textwidth]{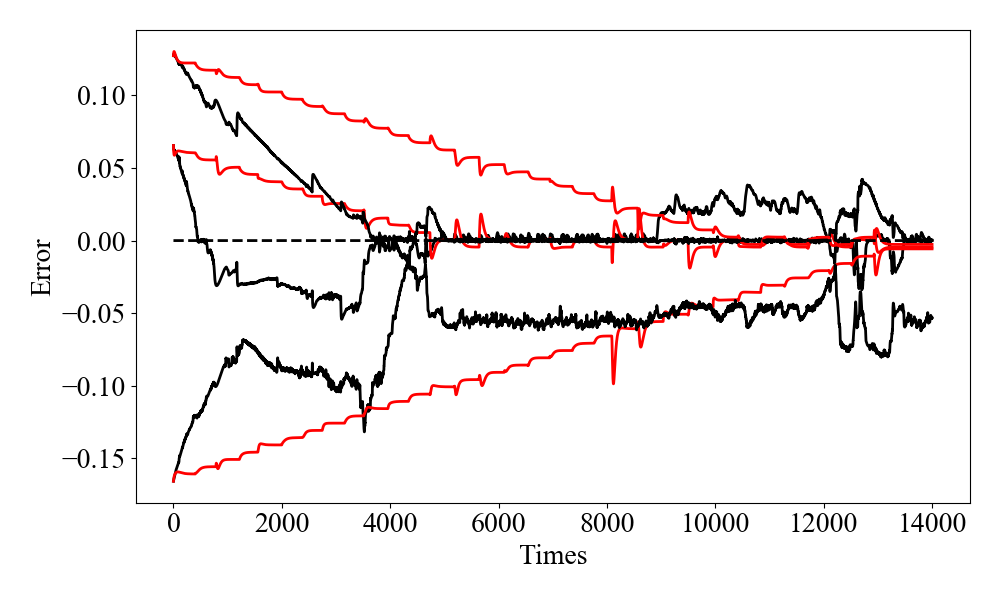}
	\caption{Trajectory evaluation near subgoals: Fixed-Frequency baseline (black line) versus the proposed Event-Driven DAMMRL progression (red line), demonstrating a significant suppression of waypoint oscillations.}
	\label{fig:comparison}
\end{figure}

\textbf{Comparison of Fixed-Frequency and Event-Triggered Control}: As illustrated in Fig. \ref{fig:comparison}, the event-driven configuration significantly reduces oscillations and improves spatial convergence compared to the fixed-frequency strategy. Fixed-frequency updates rely on rigid time intervals, often outpacing the robot's physical execution due to actuator limits and system nonlinearities. This mismatch causes premature inverse kinematics (IK) updates, which amplify residual errors and induce micro-vibrations. In contrast, the event-driven approach synchronizes commands with the robot's actual progress, triggering updates only when specific spatial thresholds are met. This acts as a spatial deadband that suppresses oscillations, reduces computational overhead, and yields smoother, highly stable joint trajectories.

\begin{figure}[htbp]
	\centering
	\begin{subfigure}[b]{0.45\textwidth}  
		\centering
		\includegraphics[width=\textwidth]{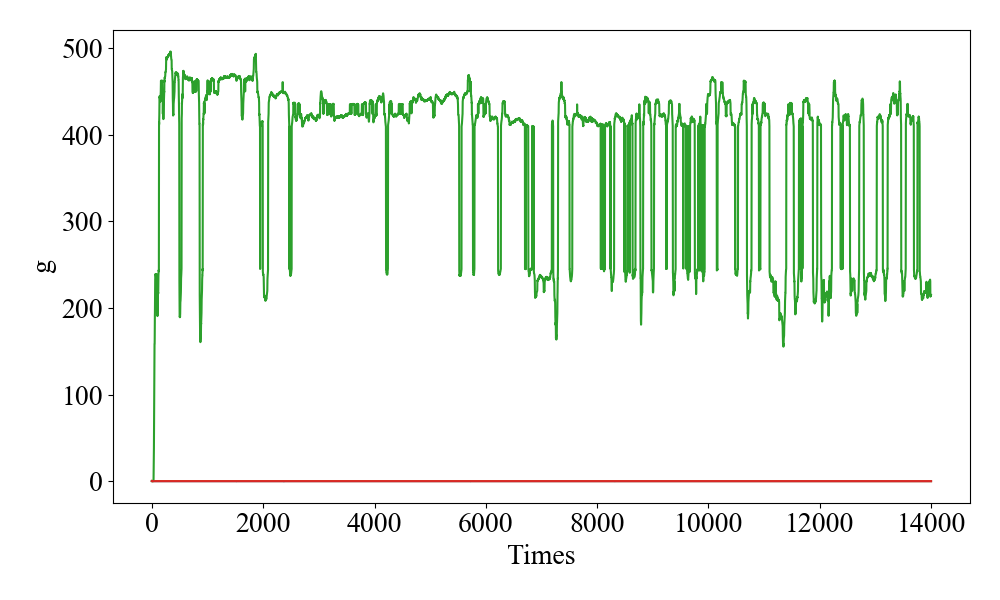}
		\caption{Human Agent input governing the primary reaching axis, measured via a physical pressure sensor.}
		\label{fig:pressure}
	\end{subfigure}
	\hspace{0.02\textwidth}
	\begin{subfigure}[b]{0.45\textwidth}
		\centering
		\includegraphics[width=\textwidth]{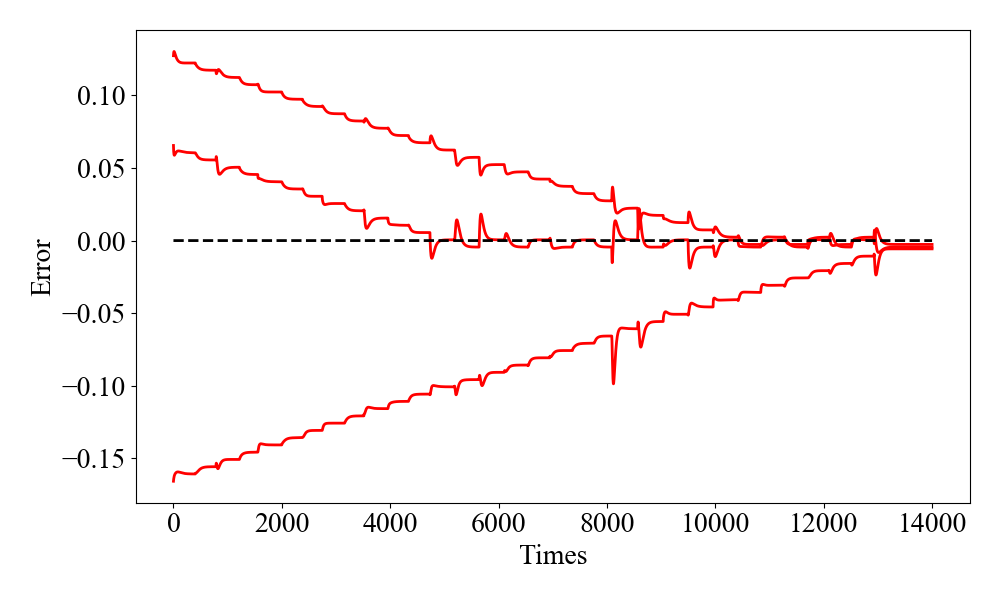}
		\caption{The spatial error convergence between the current end-effector position and the target point.}
		\label{fig:multi}
	\end{subfigure}
	\caption{Semi-Virtual (S2) experimental validation of the axial decomposition policy and event-triggered progression.}
	\label{fig:system_overview}
\end{figure}

\textbf{Human-Sim interaction Experiment}: Building on the event triggered control, we conducted a semi-virtual (Human--sim) experiment where real participants interacted with MuJoCo-simulated robot. The volunteers controlled one dimension of the robot (the up/down about z-axis) using a pressure sensor. The pressure sensor values are presented in Fig. \ref{fig:pressure}, while the positional error between the robot's end-effector and the target is illustrated in Fig. \ref{fig:multi}. These results demonstrates that the proposed event-triggered control algorithm enables the robot to stably reach the target point under control.

\begin{figure}[t]
	\centering
	
	\begin{subfigure}{0.48\linewidth}
		\centering
		\includegraphics[width=\linewidth]{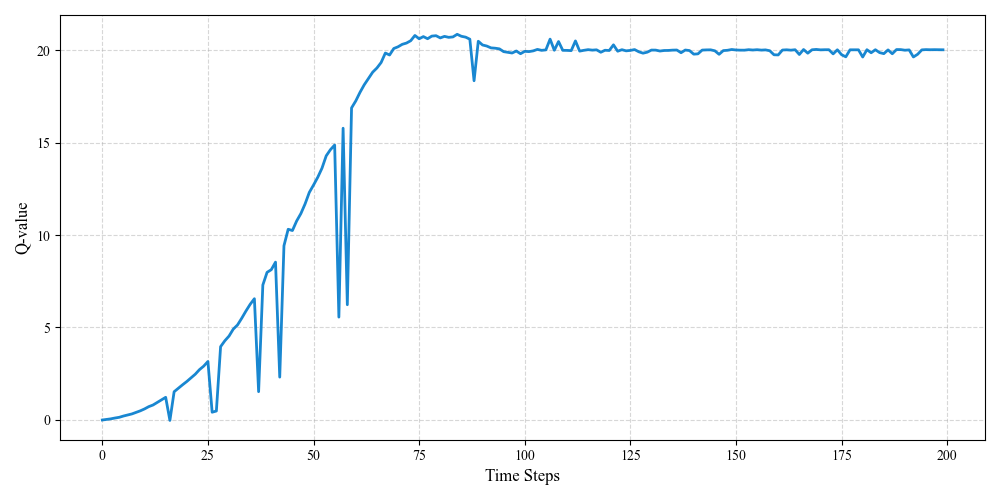}
	\end{subfigure}
	\hfill
	\begin{subfigure}{0.48\linewidth}
		\centering
		\includegraphics[width=\linewidth]{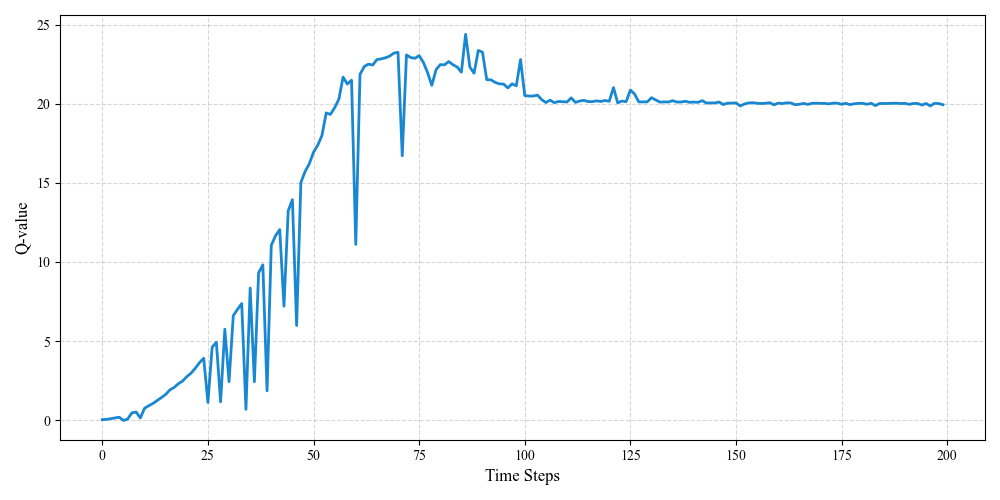}
	\end{subfigure}
	
	\vspace{0.5em}
	
	\begin{subfigure}{0.48\linewidth}
		\centering
		\includegraphics[width=\linewidth]{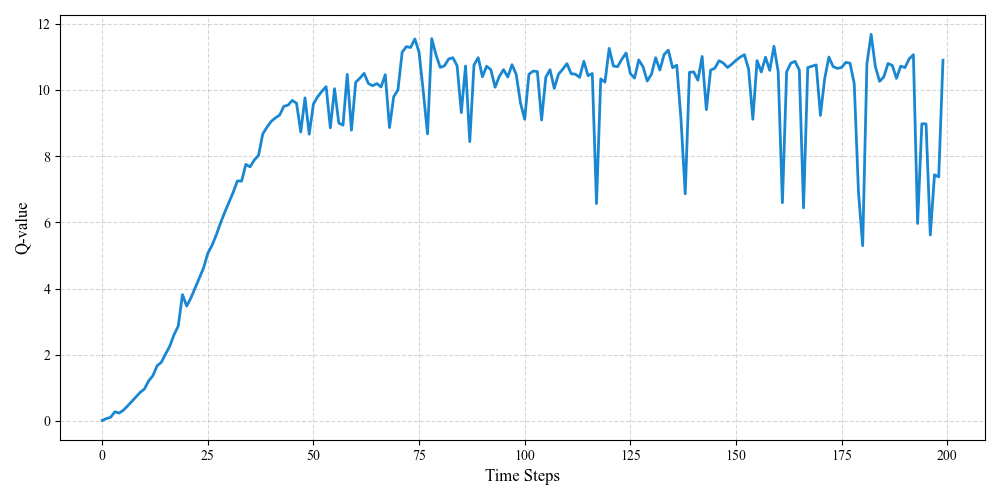}
	\end{subfigure}
	\hfill
	\begin{subfigure}{0.48\linewidth}
		\centering
		\includegraphics[width=\linewidth]{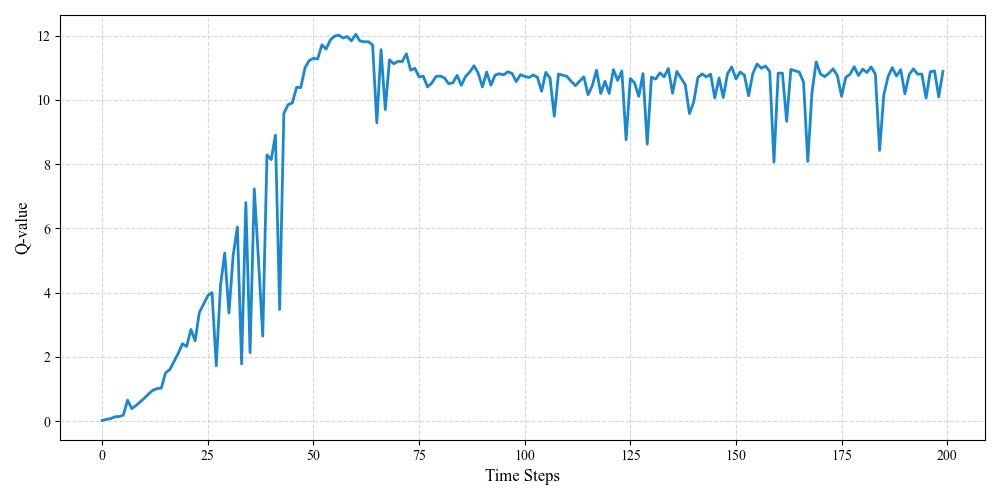}
	\end{subfigure}
	
	\caption{Training curves demonstrating the convergence of both models during the DAMMRL experiment. The top row shows convergence under Reward 1 (emphasizing spatial accuracy), while the bottom row shows convergence under Reward 2 (balancing speed and accuracy).}
	\label{fig:q-value}
\end{figure}

\begin{figure}
	\centering
	\includegraphics[width=1\linewidth]{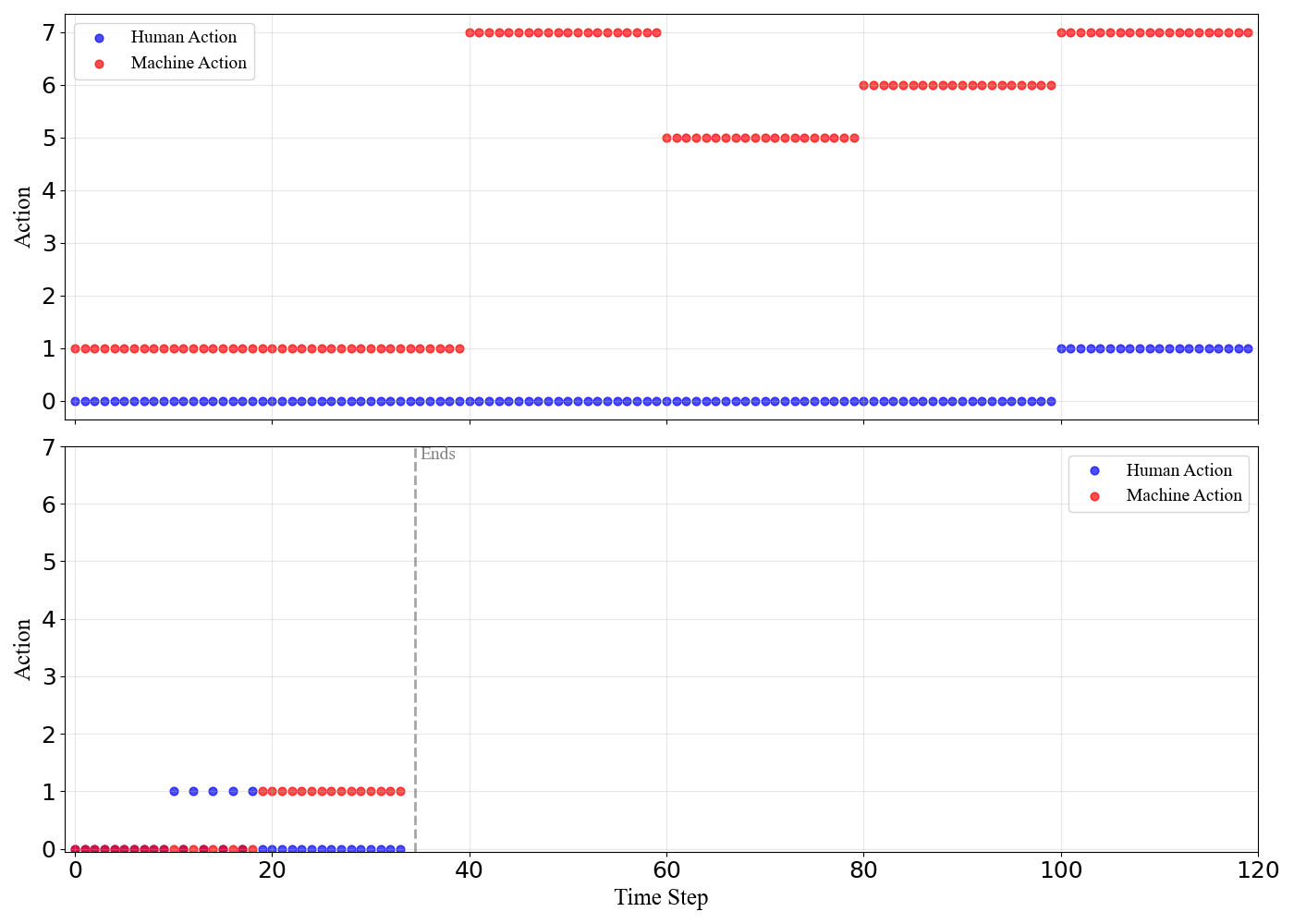}
	\caption{Comparison of the learned step-size modulation strategies and planned step counts. Top: Under Reward 1, the agent selects smaller step magnitudes near the target to prioritize spatial accuracy. Bottom: Under Reward 2, the agent opts for larger step sizes to accelerate target convergence.}
	\label{fig:action_compare}
\end{figure}

\textbf{Comparative Analysis of Convergence and Reward Effects}: In the proposed Dual Agent Multiple Model Reinforcement Learning (DAMMRL) experiment, both models have converged during training, as shown in Fig. \ref{fig:q-value}. Furthermore, as illustrated in Fig.~\ref{fig:action_compare}, the reward formulation significantly influences the learned step-size modulation strategy. Under Reward 1, which exclusively emphasizes spatial accuracy, the agent progressively adopts high-precision small step magnitudes as the end-effector approaches the target point, thereby minimizing terminal positional error. In contrast, when trained with Reward 2, which explicitly balances speed and accuracy, the agent exhibits a tendency to select comparatively larger step sizes, particularly in the mid-to-late stages of motion, in order to accelerate target convergence. This behavioral difference reflects the agent’s adaptive regulation of axial step magnitudes in accordance with the specified optimization objective.

In conclusion, the integration of event-triggered mechanisms and the DAMMRL architecture yields a highly robust and adaptable control framework. By eliminating the physical execution mismatches inherent in fixed-frequency updates, the system achieves spatial stability, which is proven crucial for the demonstrated human-in-loop interactions. Furthermore, the dual-agent RL framework successfully captures the trade-off between precision and speed, offering flexible step-size modulation based on tailored reward structures.

\section{Discussion and Limitations}
The proposed axial decomposition policy effectively reduces the human's cognitive burden to binary directional decisions coupled with an admission sphere selection. However, it currently assumes a decoupled task frame where the primary $x$-axis remains strictly aligned with the reaching direction. Complex, highly curved reaching paths could be addressed in future iterations by sequencing dynamic local task frames or spline-based subgoals. 

Furthermore, while the quantised, finite-model set $\mathcal{M}$ simplifies the DAMMRL co-adaptation process and prevents unpredictable online kinematic shifts, it may occasionally miss edge-case user preferences; introducing bounded online interpolation presents a natural algorithmic extension. Finally, while the Semi-Virtual (S2) and Real (S3) setups successfully validated the system with healthy participants, extensive clinical validation on neurologically impaired patient cohorts remains a critical next step.

\section{Conclusion}
This study introduces an event-driven, axial decomposition policy that seamlessly couples binary human intent with autonomous orthogonal corrections on a custom 6-DoF upper-limb rehabilitation robot. By governing task progression via admission spheres rather than traditional fixed-time sampling, the architecture successfully mitigates the waypoint oscillations frequently caused by variable execution times in inverse-dynamics mappings. Furthermore, the proposed Dual Agent Multiple Model Reinforcement Learning (DAMMRL) framework effectively addresses inter-individual variability by discretely matching the human's chosen speed--accuracy trade-off with the robot's dynamically sized Cartesian micro-steps. Validated through a staged training curriculum transitioning from MuJoCo simulations to physical hardware, the system demonstrates superior spatial precision, temporal efficiency, and task success rates compared to conventional fixed-frequency shared control approaches.

\bibliographystyle{splncs04}
\bibliography{reference}
\end{document}